\documentclass[journal]{IEEEtran}

\ifCLASSINFOpdf
\else
   \usepackage[dvips]{graphicx}
\fi
\usepackage{url}

\usepackage{graphicx}

\usepackage{subfigure}  

\usepackage{color}
\usepackage{multirow}
\usepackage{amsmath}
\usepackage{booktabs}
\usepackage{float}
\usepackage{bm}
\usepackage{wrapfig}
\usepackage{mathrsfs}
\usepackage{bbding}
\usepackage{amssymb}
\usepackage{caption}

\usepackage{amsfonts}       
\usepackage{nicefrac}       
\usepackage{microtype}      
\usepackage{xcolor}

\usepackage{array} 
\usepackage{pifont}

\usepackage{bbding}

\newcommand{\ie}{\textit{i}.\textit{e}., }
\newcommand{\eg}{\textit{e}.\textit{g}., }
\newcommand{\etc}{\textit{etc}}

\begin{document}

\title{Dual-Augmented Transformer Network for Weakly Supervised Semantic Segmentation}

\author{Jingliang Deng, Zonghan Li

\thanks{Jing. D and Zong. L are with the School of Software Engineering, South China University of Technology.}
}

\markboth{Journal of \LaTeX\ Class Files, Vol. 14, No. 8, April 2023}
{Shell \MakeLowercase{\textit{et al.}}: Bare Demo of IEEEtran.cls for IEEE Journals}
\maketitle

\begin{abstract}
Weakly supervised semantic segmentation (WSSS), a fundamental computer vision task, which aims to segment out the object within only class-level labels. The traditional methods adopt the CNN-based network and utilize the class activation map (CAM) strategy to discover the object regions. However, such methods only focus on the most discriminative region of the object, resulting in incomplete segmentation. An alternative is to explore vision transformers (ViT) to encode the image to acquire the global semantic information. Yet, the lack of transductive bias to objects is a flaw of ViT. In this paper, we explore the dual-augmented transformer network with self-regularization constraints for WSSS. Specifically, we propose a dual network with both CNN-based and transformer networks for mutually complementary learning, where both networks augment the final output for enhancement. Massive systemic evaluations on the challenging PASCAL VOC 2012 benchmark demonstrate the effectiveness of our method, outperforming previous state-of-the-art methods.

\end{abstract}


\IEEEpeerreviewmaketitle

\section{Introduction}

\IEEEPARstart{I}{mage} segmentation~\cite{chen2014semantic,lin2017refinenet} and detection~\cite{tian2019fcos,su2022epnet} are fundamental tasks in computer vision, which can help us analyze the essence of an image. However, collecting such pixel-wise labels is time-costly, and thus, recent research interest has shifted to weakly-supervised image segmentation (WSSS)~\cite{wang2020self,su2021context,huo2022dual} and weakly-supervised object localization (WSOL)~\cite{su2022self, cao2023semantic}. In this paper, we focus on WSSS task, in which the mainstream is first to mine the potential object regions through class activation maps (CAMs)~\cite{zhou2016learning} to obtain the pseudo-masks. The fully-supervised image segmentation network~\cite{chen2017deeplab,chen2018encoder,sun2023denet} is utilized in the second stage for end-to-end training.

\begin{figure}
	\begin{center}
		\centering
		\includegraphics[width=3.3in]{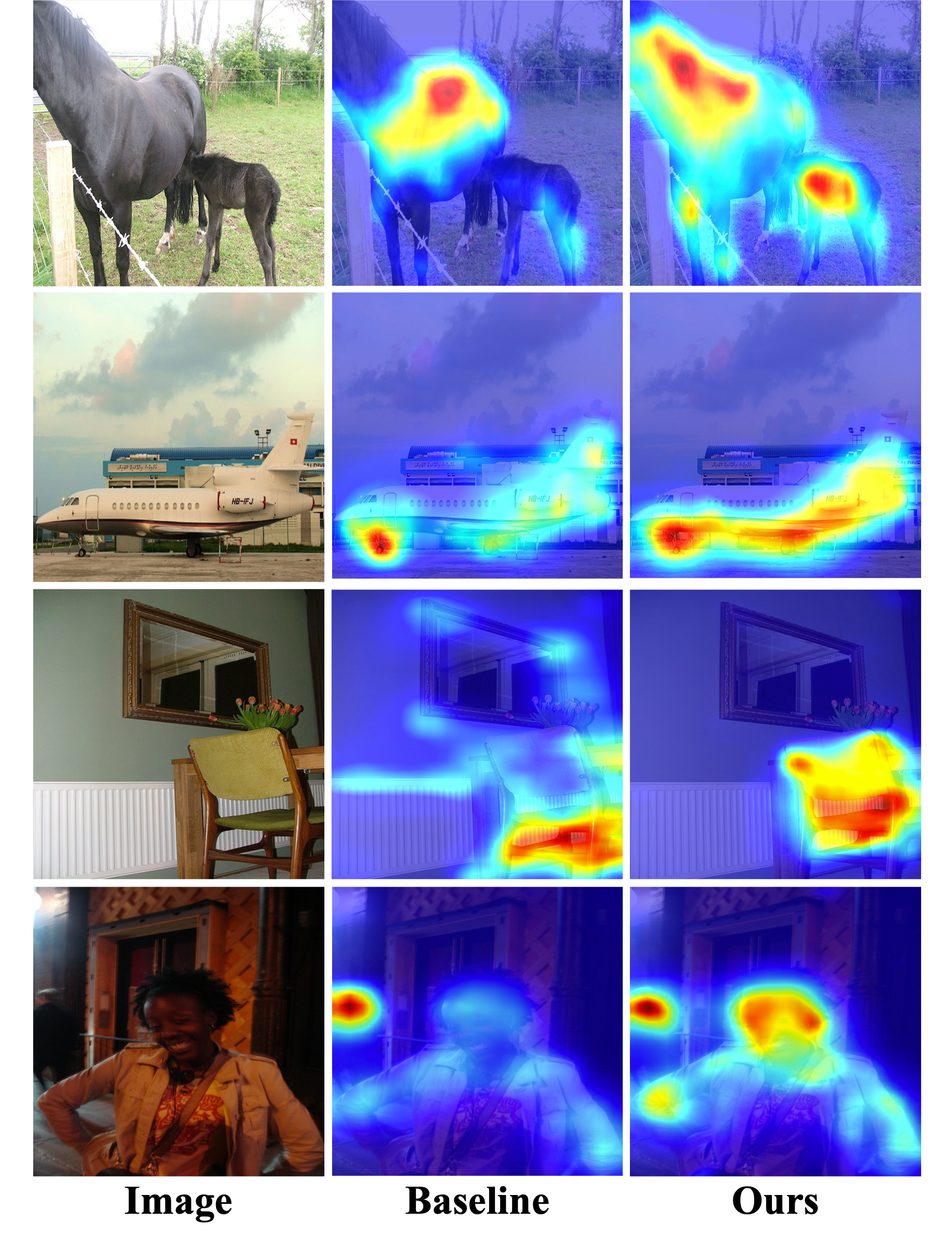}
	\end{center}
	\caption{An illustrative example of the visual effect of different methods. Our method can produce better activation maps of the object covering more regions than the baseline model, which is beneficial to yield accurate pseudo masks.}
	\label{fig1}
	\vspace{-8pt}
\end{figure}

However, the previous methods usually adopt CNN-based networks, which may yield locally activated class maps, as shown in Fig~\ref{fig1}(middle). This is because CNN uses local receptive field convolutional operations for image encoding, resulting in poor pseudo masks for the following supervised training.
To this end, some papers tried to utilize vision transformers (ViT)~\cite{dosovitskiy2020image} as image encoders in this task. Yet, ViT lacks inductive bias~\cite{goyal2020inductive}, which is less sensitive to the existence of decisive features but pays attention to the global location of features. This may cause cross-regional activation.

Motivated by the CNN-based and augmented transformer network~\cite{peng2021conformer}, we find that the class activation map output from the CNN-based and ViT networks can be mutually complementary learning.
Therefore, in this paper, we introduce a dual-augmented transformer network for weakly supervised semantic segmentation, whose activation map can be referred to Fig~\ref{fig1}(the third column). Specifically, given the input RGB image, it will be fed into the dual CNN-based (\ie ResNet50~\cite{he2016deep}) network and transformer network (\ie ViT~\cite{dosovitskiy2020image}).
More importantly, we experimentally explore that the output distributions from different networks are inconsistent. Therefore, we further introduce a self-regularization loss to constrain both the features and calibrate the features, focusing on more foreground object regions without being over-activated or partially activated. The key idea behind this is that we hope that the transformer and CNN can complement each other. This constraint can be leveraged to learn the global consistency of entire object representations.

The contributions of our approach are listed as follows: 

\begin{itemize}
	\item We take the early step to explore the usefulness of the vision transformer in the WSSS task and experimentally analyze the different output distributions from different architectures.
	\item We propose a dual-augmented transformer network for WSSS, which combines both the global long-range dependency and local inductive bias from a transformer and convolutional network. The self-regularization mechanism constrains the activation maps focusing on more foreground objects.
	\item Massive systemic evaluations conducted on the public benchmark demonstrate the superiority of the method, and it can outperform some previous state-of-the-art methods.
\end{itemize}

\begin{figure*}[t]
		\begin{center}
			\centering
			\includegraphics[width=7.2in]{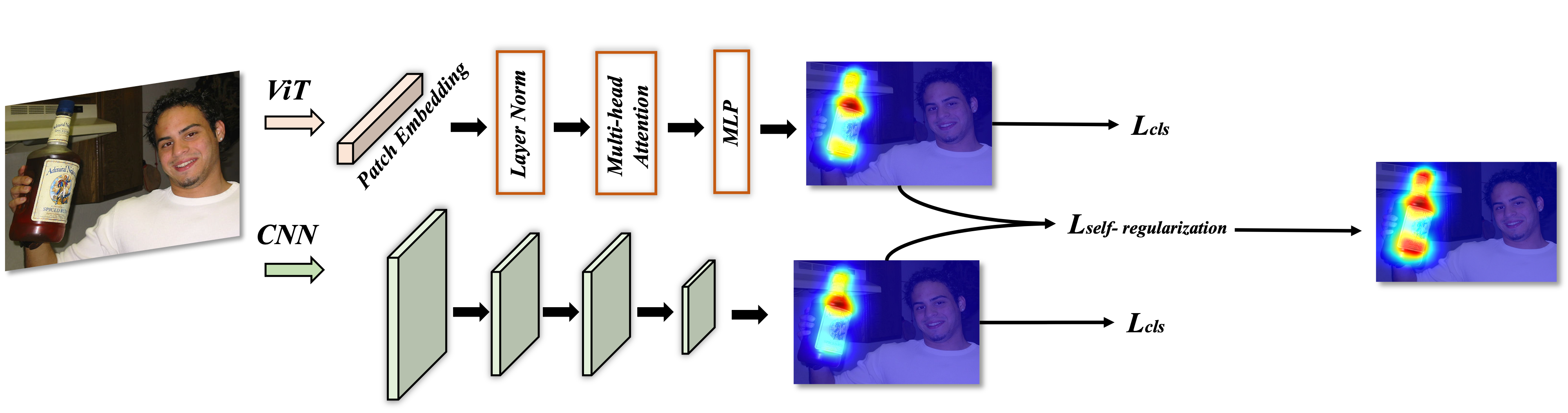}
		\end{center}
		\caption{The overall architecture of our proposed dual-augmented transformer network for weakly supervised semantic segmentation. The input image is encoded by both the CNN-based and ViT networks, where both networks augment the final output for enhancement.}
		\label{fig2}
\end{figure*}

\section{Related Work}
\label{sec:guidelines}

\textbf{Weakly Supervised Semantic Segmentation.} 
Weakly Supervised Semantic Segmentation (WSSS) aims to only utilize class-wise labels to segment out the objects, which is based on the class activation map (CAM) technique~\cite{zhou2016learning}.
Later, since the CAM strategy failed to mine the entire cues of the objects, many different improvements have been proposed. In the early time, some of the erasing methods~\cite{wei2017object,hou2018self} are proposed to iteratively excavate the object regions. Then, IRNet~\cite{ahn2019weakly} and AffinityNet~\cite{ahn2018learning} attempt to mine the objects using pixel-affinity mining strategies and edge information. SEAM~\cite{wang2020self} introduces a self-supervised equivariant attention mechanism for WSSS. Sun $\emph{et~al.}$~\cite{sun2020mining} later proposes a cross-region mining network via encoding the co-occurrent object images. Besides, Su $\emph{et~al.}$~\cite{su2021context} introduces a context decoupling augmentation strategy to mine the potential object step-by-step. 
Besides, some of the state-of-the-art methods~\cite{lee2019ficklenet,huang2018weakly} adopt saliency maps as auxiliary information to enhance performance. CONTA~\cite{zhang2020causal} explores causal intervention mechanism to reduce the bias in WSSS.
However, in general, all the above-mentioned methods use CNN-based networks to encode the images, which hinders the global receptive field in vision analysis.

\vspace{1ex}

\textbf{Vision Transformer.} 
Vision transformer (ViT)~\cite{dosovitskiy2020image} is first proposed to process the image data, which is similar to sequential data in the natural language processing domain~\cite{vaswani2017attention}. Because of its strong parallelism and globality, ViT has become more and more popular in the vision area.
Su $\emph{et~al.}$~\cite{su2023unified} explores a vision transformer to model the global long-range dependencies for group-based segmentation tasks.
Gao $\emph{et~al.}$~\cite{gao2021ts} and Cao $\emph{et~al.}$~\cite{cao2023semantic} introduce to use the transformer in weakly supervised localization, which inspired some follow-up studies focusing on this task. Moreover, vision transformers are proved to be efficient and useful in low-level~\cite{liang2021swinir}, inpainting~\cite{liu2021fuseformer} and AIGC~\cite{rombach2022high}, \etc.

\vspace{1ex}

\textbf{Vision Constraint.} 
Vision constraint is widely used in many tasks for enhancement. For instance, Su $\emph{et~al.}$~\cite{su2022self} explores self-constraint loss for unsupervised object localization. Besides, Liu $\emph{et~al.}$~\cite{liu2020weakly} used the maximum bipartite graph matching constraint strategy for weakly supervised semantic segmentation. Furthermore, some works utilize high-level constraints for pose transformation~\cite{su2022general} and image matching~\cite{sarlin2020superglue}, \etc.

\section{Method}
In this section, we will introduce the details of our framework including the proposed dual-augmented network and the supervised segmentation training process.

\begin{figure*}[t]
		\begin{center}
			\centering
			\includegraphics[width=6.0in]{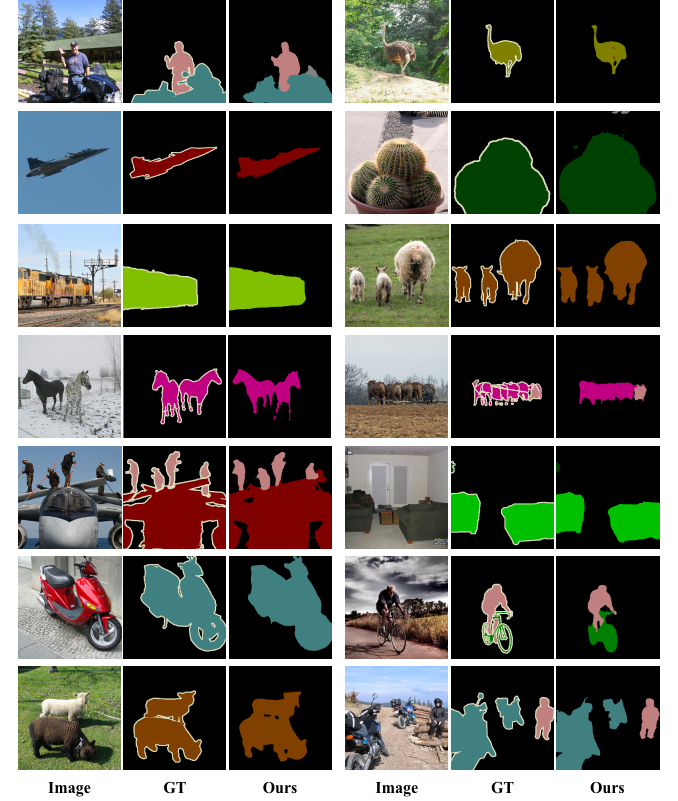}
		\end{center}
		\caption{Some qualitative results on the PASCAL VOC 2012 val set. It can be seen that our predictions are close to the ground-truth only using class labels to train the network.}
		\label{fig3}
\end{figure*}

\subsection{Network}
Fig~\ref{fig2} shows the overall network architecture of the proposed dual-augmented transformer network. Specifically, given an input image $X$, it will be fed into two different encoders. For the CNN-based encoder, we adopt ResNet50~\cite{he2016deep} as the encoder and finally get the feature map $\mathcal{M}_C$ as follows:

\begin{equation}
\mathcal{M}_{C} = f_{CNN}(X),
\tag{1}
\label{eq1}
\end{equation}
where $f_{CNN}$ denotes the CNN-based encoder.

However, as we mentioned before, the output from the CNN-based network only focuses on the most discriminative part of the objects, making the class activation maps locally activated. Therefore, the input image is also encoded by the transformer network (\eg ViT~\cite{dosovitskiy2020image}) to capture the global semantic information. Concretely, similar to the ViT input, we first reshape the 2D grid image into a sequential embedding for ViT. Then, for each transformer block, we compute the forward training as follows:

\begin{equation}
\begin{split}
X_n^{'} &= \text{MSA}(\text{LN}(X_n)) + X_n, \\
X_n^{'} &= \text{MLP}(\text{LN}(X_n^{'})) + X_n^{'}, \\
X_n^{'} &= \text{LN}(X_n^{'})
\end{split}
\tag{2}
\label{eq2}
\end{equation}
where $X_n$ indicates the $n^{th}$ layer input of the transformer block. To save computing resources, we adopt vit-tiny~\cite{dosovitskiy2020image} as the transformer backbone.
Ultimately, we reshape the embedding from the transformer back into the 2D-grid shape feature map by using a standard 1 $\times$ 1 convolutional layer as follows:

\begin{equation}
\mathcal{M}_{C} = \textit{Conv1x1}(X_n),
\tag{3}
\label{eq3}
\end{equation}

For both dual backbones, we generate the class-activation map by adopting the multi-class classification loss as follow for gradient backward propagation training:

\begin{equation}
\begin{split}
\mathcal{L}_{cls} = &-\sum_{c=1}^{C} y_c log(o_c) \\
& + (1 - y_c) log(1- o_c),
\end{split}
\tag{4}
\label{eq4}
\end{equation}
where $o_c$ is the model’s prediction for the $c$-th class, $y_c$ is the image-level label for the $c$-th class and $C$ is the total number of foreground classes.

However, by simply applying the multi-class cross-entropy loss is not strong enough to restrict the different out distributions from both the different encoders. To tackle this drawback, we further introduce a constraint self-regularization loss to calibrate the CAM from the transformer backbone and the CNN-base network computed as follows:

\begin{equation}
\mathcal{L}_{self-regularization} = \text{Smooth}_{L1}|| \mathcal{M}_{T} - \mathcal{M}_{C} ||_{1}, \tag{5}
\label{eq5}
\end{equation}
where $\mathcal{M}_{T}$ and $\mathcal{M}_{C}$ are the CAMs output form the transformer and CNN, respectively. The smooth L1 loss guarantees both the activation maps can be optimized softly and close to each other.

In general, we train our network in an end-to-end manner by combing all the losses as follows:

\begin{equation}
\mathcal{L}_{total} = \mathcal{L}_{cls_{transformer}} + \mathcal{L}_{cls_{CNN}} + \mathcal{L}_{self-regularization}. \tag{6}
\label{eq6}
\end{equation}

Note that during the inference phase, we only use the transformer encoder for weakly supervised semantic segmentation.

\subsection{Supervised Segmentation}
Similar to the previous WSSS works~\cite{wang2020self,ahn2018learning}, after obtaining the CAM, we adopt the state-of-the-art post-processing techniques (\eg CAM + random walk~\cite{ahn2018learning}) to generate the pseudo-masks to train the fully-supervised encoder-decoder segmentation network. For fair comparisons, we use Deeplab~\cite{chen2014semantic} to train the segmentation network using the standard pixel-wise loss as follows:

\begin{equation}
\begin{split}
\mathcal{L}_{cls} = &-\frac{1}{HW} \sum_{i=1}^{H} \sum_{i=1}^{w} Y(i,j) log(P(i,j)) \\
& + (1 - Y(i,j)) log(1- P(i,j)),
\end{split}
\tag{7}
\label{eq7}
\end{equation}
where $H$ and $W$ are the height and width of the images, $G(i,j)$ and $P(i,j)$ denote the ground-truth labels and prediction.

\section{Experiments}

\subsection{Dataset and Metrics}
Following the previous works~\cite{wang2020self,su2021context} strictly, we conduct experiments on the PASCAL VOC~\cite{everingham2010pascal} dataset, which contains 1464 images for training, 1449 for validation and 1456 for testing. Following the common experimental protocol for semantic segmentation, we utilize additional annotations from SBD~\cite{hariharan2011semantic} to augment the training set with 10582 images. Moreover, we use the standard mean Intersection-over-Union (mIoU) as the evaluation metric for all experiments.

\subsection{Implementation}
For fair comparisons, the input size is set to 224 $\times$ 224 in our network. We use ViT~\cite{dosovitskiy2020image} as our transformer encoder and ResNet50~\cite{he2016deep} as our CNN-based network. We use Adam optimizer with $\epsilon$ = 1e-8, $\beta_1$ = 0.9 and $\beta_2$ = 0.99 and weight decay of 5e-4 to train our dual-augmented network. Random rotation and flip augmentation strategies are adopted. The network is trained on 4 TITAN-Xp GPUs with batch size 32 for 80 epochs. After training and acquiring the pseudo-mask labels, we use Deeplab~\cite{chen2014semantic} as~\cite{wang2020self,su2021context} to train a fully-supervised network to yield the final predictions.

\begin{table}[]
	\begin{center}
		\scalebox{0.9}{
			\begin{tabular}{cccc}
				\toprule  
				\toprule  
				Transformer  & CNN & $\mathcal{L}_{self-regularization}$  & mIoU (\%)\\
				\midrule  
				$\checkmark$&  &  & 30.3 \\
				&  $\checkmark$& & 48.3 \\
				$\checkmark$&  $\checkmark$ &  & 55.2 \\
				$\checkmark$&  $\checkmark$&$\checkmark$  & \textbf{60.4} \\
				\bottomrule 
		\end{tabular}}
	\end{center}\caption{Each component's accuracy in our proposed network on the PASCAL VOC 2012 training set in mIoU.}\label{table1}
\end{table}

\begin{table}[]
	\begin{center}
		\scalebox{0.9}{
			\begin{tabular}{ccc|ccc}
				\toprule  
				\toprule  
				CAM~\cite{zhou2016learning} &  + RW & + RW + dCRF & Ours &  + RW &  + Rw + dCRF \\
				\midrule  
				48.0 & 58.1 & 59.7 & 60.4 & 66.9 & 68.3 \\
				\bottomrule 
		\end{tabular}}
	\end{center}\caption{Accuracy of synthesized pseudo labels in mIoU on the PASCAL VOC 2012 training set.}\label{table3}
\end{table}

\begin{table}[]
	\begin{center}
		\scalebox{1.2}{
			\begin{tabular}{ccc|cc}
				\toprule  
				\toprule  
				Methods & Backbone &  Saliency & $\mathit{val}$ & $\mathit{test}$\\
				\midrule  
				CCNN~\cite{pathak2015constrained} & VGG16 & -& 35.3 & 35.6\\
				SEC~\cite{kolesnikov2016seed} & VGG16 & -&50.7& 51.1\\
				STC~\cite{wei2016stc} & VGG16 & \checkmark&49.8 & 51.2\\
				AdvEra~\cite{wei2017object}& VGG16 & \checkmark & 55.0 & 55.7\\
				DCSP~\cite{chaudhry2017discovering}& ResNet101 & \checkmark & 60.8 & 61.9\\
				MDC~\cite{wei2018revisiting}& VGG16 & \checkmark & 60.4 & 60.8\\
				MCOF~\cite{wang2018weakly}& ResNet101 & \checkmark & 60.3 & 61.2\\
				DSRG~\cite{huang2018weakly}& ResNet101 & \checkmark & 61.4 & 63.2\\
				AffinityNet~\cite{ahn2018learning}& ResNet-38 & - & 61.7 & 63.7\\
				IRNet~\cite{ahn2019weakly}& ResNet50 & - & 63.5 & 64.8\\
				FickleNet~\cite{lee2019ficklenet}& ResNet101 & \checkmark & 64.9 & 65.3\\
				SEAM~\cite{wang2020self}& ResNet38 & - & 64.5 & 65.7\\
				ICD~\cite{fan2020learning} & ResNet101 & - & 64.1 & 64.3 \\
				CDA~\cite{su2021context} & ResNet50 & - & 66.1 & 66.8 \\
				\midrule  
				Ours & Transformer & - & \textbf{68.9} & \textbf{68.7} \\
				\bottomrule 
		\end{tabular}}
	\end{center}\caption{Performance comparisons with other state-of-the-art WSSS methods on PASCAL VOC 2012 dataset.}\label{table2}
\end{table}

\subsection{Ablation Studies}
We first explore the usefulness of the proposed components of our framework. As shown in Table1~\ref{table1}, we can observe that using only the CNN-based or the transformer encoder can not achieve competitive performance. When we utilize the dual-augmented transformer network, we can yield satisfactory performance and boost the performance by a large margin. This validates the effectiveness of the proposed network.

Furthermore, we analyse the quality of the synthesized pseudo label, as shown in tab~\ref{table3}. We can also observe that by applying the same post-process strategy (\eg Random Walk~\cite{ahn2018learning} and DenseCrf~\cite{krahenbuhl2011efficient}), our proposed framework can achieve better performance. This demonstrates that we can produce more accurate pseudo-masks for the supervised segmentation network.

\subsection{Comparison with State-of-the-arts}
In this section, we compare our proposed network with the recent state-of-the-art WSSS methods. As shown in Table~\ref{table2}, our network can achieve the new state-of-the-art performance. It is worth mentioning that our method can even outperform some of the methods that use additional saliency maps. All these systematic evaluations reveal the superiority of our network. And the proposed network does not require any complex human-designed techniques for training. In other words, if computing resources are sufficient, one can replace stronger CNN-based or transformer encoders for this task.

Finally, we visualize some qualitative results in Fig~\ref{fig3}. As can be seen, we can produce high-quality segmentation masks by merely using class-level labels, which are competitive.

\section{Conclusion}
In this paper, we experimentally explore the uselessness of vision transformer in weakly supervised semantic segmentation (WSSS) task, and analyze the different output distributions between CNN-based methods and ViT. We introduce the dual-augmented transformer network within the self-regularization loss to mine the global representations of the objects. Extensive experimental results show that our proposed network can achieve the new state-of-the-art performance.

\bibliographystyle{IEEEtran}
\bibliography{mybib}

\end{document}